%% file: main.tex
\documentclass[10pt,twocolumn,letterpaper]{article}
\usepackage{algorithm}
\usepackage{algorithmic}
\usepackage{booktabs}
\usepackage{graphicx}
\usepackage{longtable}
\usepackage{tcolorbox}
\usepackage{stackengine}
\usepackage[T1]{fontenc}
\usepackage[accsupp]{axessibility}
\usepackage[pagenumbers]{cvpr} 

\usepackage{enumitem}
\usepackage{marvosym}

\definecolor{cvprblue}{rgb}{0.21,0.49,0.74}
\usepackage[pagebackref,breaklinks,colorlinks,citecolor=cvprblue]{hyperref}

\def\confName{CVPR}
\def\confYear{2024}

\title{On the Robustness of Language Guidance for Low-Level Vision Tasks: \\
Findings from Depth Estimation}

\author{
Agneet Chatterjee\textsuperscript{$\diamondsuit$} \qquad Tejas Gokhale\textsuperscript{$\spadesuit$} \qquad Chitta Baral\textsuperscript{$\diamondsuit$} \qquad Yezhou Yang\textsuperscript{$\diamondsuit$} \\
\textsuperscript{$\diamondsuit$}Arizona State University \qquad \textsuperscript{$\spadesuit$}University of Maryland, Baltimore County\\
{\small \tt agneet@asu.edu, gokhale@umbc.edu, chitta@asu.edu, yz.yang@asu.edu} 
}

\begin{document}
\maketitle
\input{sections/0_abstract}    
\input{sections/1_intro}
\input{sections/2_related_work}

\input{sections/3_language_guided_depth_estimation}
\input{sections/4_experiments}

\input{sections/5_robustness}
\input{sections/6_conclusion}
\input{sections/7_acknowledgement}
{
    \small
    \bibliographystyle{ieeenat_fullname}
    \bibliography{ref}
}

\input{suppl}
\end{document}

%% file: sections/0_abstract.tex
\begin{abstract}
Recent advances in monocular depth estimation have been made by incorporating natural language as additional guidance. 
Although yielding impressive results, the impact of the language prior, particularly in terms of generalization and robustness, remains unexplored. 
In this paper, we address this gap by quantifying the impact of this prior and introduce methods to benchmark its effectiveness across various settings. 
We generate "low-level" sentences that convey object-centric,  three-dimensional spatial relationships, incorporate them as additional language priors and evaluate their downstream impact on depth estimation. 
Our key finding is that current language-guided depth estimators perform optimally only with scene-level descriptions and counter-intuitively fare worse with low level descriptions. 
Despite leveraging additional data, these methods are not robust to directed adversarial attacks and decline in performance with an increase in distribution shift. 
Finally, to provide a foundation for future research, we identify points of failures and offer insights to better understand these shortcomings. 
With an increasing number of methods using language for depth estimation, our findings highlight the opportunities and pitfalls that require careful consideration for effective deployment in real-world settings. \footnote{Project Page : \url{https://agneetchatterjee.com/robustness_depth_lang/}}

\end{abstract}

%% file: sections/1_intro.tex
\section{Introduction}
\label{sec:intro}

\begin{figure*}[t]
    \centering
    \includegraphics[width=\linewidth]{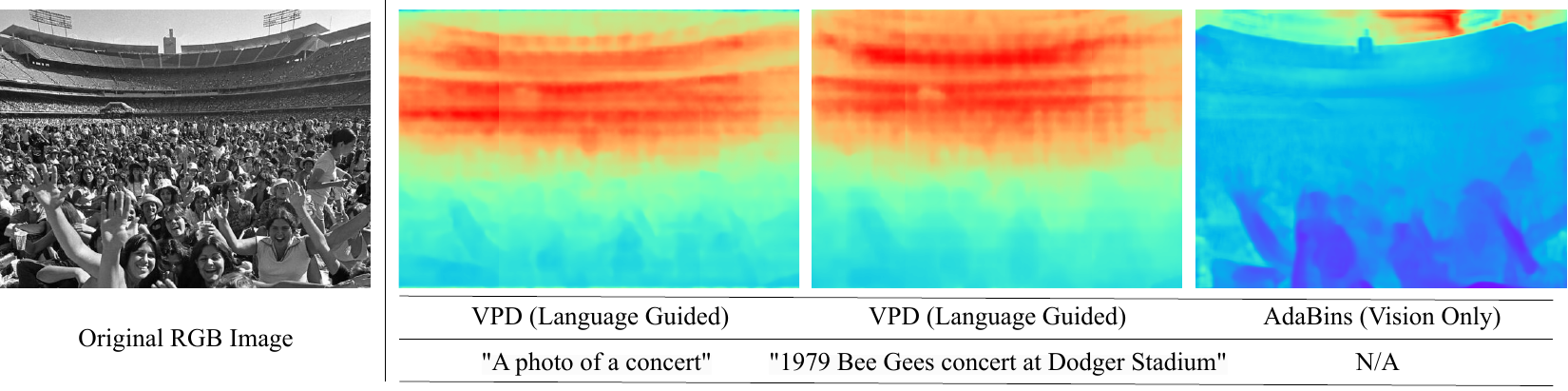}
    \caption{We investigate the efficacy of language guidance for depth estimation by evaluating the robustness, generalization, and spurious biases associated with this approach, comparing it alongside traditional vision-only methods. 
    Shown here is a visual comparison of the depth estimation results between VPD (with additional knowledge) and AdaBins \cite{Farooq_Bhat_2021} on an out-of-domain outdoor scene.
    }%
    \label{fig:teaser}
\end{figure*}

\begin{figure*}
    \centering 
    \includegraphics[width=\linewidth]{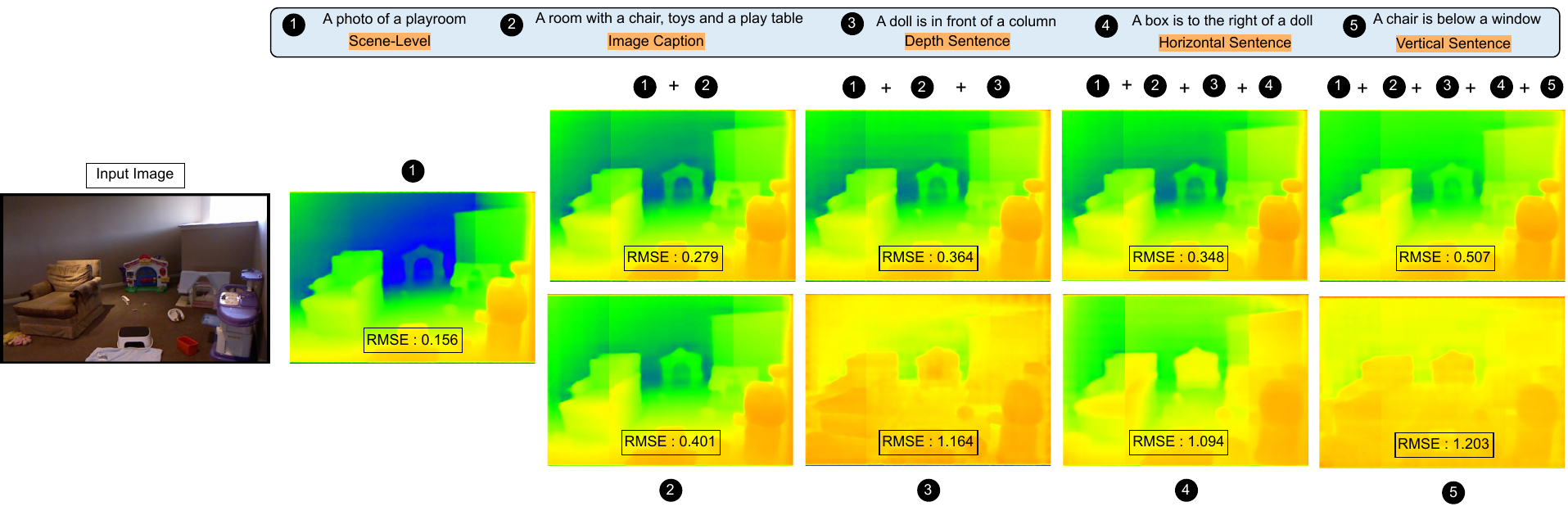}
    \caption{An illustration of depth maps generated by language-guided depth estimation methods such as VPD (\textbf{zero-shot}) when prompted with various sentence inputs that we use as part of our study. 
    The first row shows the effect of progressively adding descriptions as input, while the second row shows depth maps generated by single sentence inputs.}%
    \label{fig:image_1}
\end{figure*}

Computational theories of visual perception have proposed hierarchies of visual understanding, such as in the work of Gestalt psychologists \cite{desolneux2004gestalt}, Barrow and Tenenbaum \cite{barrow1981computational}, and others.
In this hierarchy, \textit{higher-level} vision tasks are aligned with semantics or human-assigned labels; for instance, recognizing scenes, detecting objects and events, answering questions and generating captions about images, or retrieving and generating images from text queries.
Breakthroughs in large-scale vision–language pretraining \cite{radford2021learning, li2022blip, wang2022image} and diffusion-based modeling techniques \cite{ramesh2021zeroshot, rombach2022highresolution} have been significantly improved the state-of-the-art in such higher-level semantic visual understanding tasks.

This success has been driven by the idea that natural language is an effective and free-form way to describe the contents of an image and that leveraging this data for learning shared representations benefits downstream tasks.

\textit{Lower-level} vision has a different perspective on image understanding and seeks to understand images in terms of geometric and physical properties of the scene such as estimating the depth (distance from the camera), surface normals (orientation relative to the camera) of each pixel in an image, or other localization and 3D reconstruction tasks.

A physics-based understanding of the scene such as the measurement or estimation of geometric and photometric properties, underlies the solutions for these inverse problems.
As such, until now, state of the art techniques \cite{ming2021deep, wang2015designing} for lower-level tasks such as depth estimation have not featured the use of natural language. Surprisingly, recent findings from VPD \cite{zhao2023unleashing}, TADP \cite{tadp} and EVP \cite{EVP} have demonstrated that language guidance can help in depth estimation, thus potentially building a bridge between low-level and high-level visual understanding.

In this paper, we seek to inspect this bridge by asking a simple question: {\bf what is the impact of the natural language prior on depth estimation?}
Our study is positioned to complement early exploration into the role of natural language for training models for low-level tasks, especially given the emerging evidence of state-of-the-art performance on tasks such as depth estimation.

We argue that in the age of large language models,  
it is essential to fully grasp the implications of using language priors for vision tasks and so we examine these methods from multiple perspectives to evaluate their complete potential.
This examination is important as depth estimation is used for many real-world and safety-critical applications such as perception in autonomous vehicles, warehouse automation, and robot task and motion planning.

We conduct a systematic evaluation to determine the significance of language-conditioning and the effect of this conditioning under various settings. 
We also benchmark the generalization capabilities and robustness of these methods. 
Specifically, we create image and language-level transformations to evaluate the true low-level understanding of these models . 
We construct natural language sentences that encode low-level object-specific spatial relationships, image captions and semantic scene descriptions using pixel-level ground truth annotations. Similarly, we perform image-level adversarial attacks implementing object-level masking, comparing vision-only and language-guided depth estimators on varying degrees of distribution shift. 

Through our experiments, we discover that existing language-guided methods work only under the constrained setting of scene-level descriptions such as \textit{``a photo of a bedroom''}, but suffer from performance degradation when the inputs describe relationships between objects such as \textit{``a TV in front of a bed''},
and are unable to adapt to intrinsic image properties. 
Furthermore, with an increase in domain shift, these methods become less robust in comparison to vision-only methods. 
We present insights to explain the current challenges of robustness faced by these models and open up avenues for future research. A visual illustration of this performance gap is presented in Figure \ref{fig:teaser}.

\smallskip\noindent Our contributions and findings are summarized below: 
\begin{itemize}
    \item We quantify the guidance provided by language for depth estimation in current methods. We find that existing approaches possess a strong scene-level bias, and become less effective at localization when low-level information is provided. We additionally offer analysis grounded in foundation models to explain these shortcomings.
    \item Through a series of supervised and zero-shot experiments, we demonstrate that existing language-conditioned models are less robust to distribution shifts than vision-only models.
    \item We develop a framework to generate natural language sentences that depict low-level 3D spatial relationships in an image by leveraging ground truth pixel-wise and segmentation annotations.

\end{itemize}

Our findings underline the importance of taking a \textit{deeper} look into the role of language in monocular depth estimation. We quantify the \textit{counter-intuitive} finding that a sentence such as "\textit{The photo of a kitchen}" leads to better depth estimation than "\textit{A knife is in front of a refrigerator}", although the latter explicitly contains depth information. 
In this paper, we provide trade-offs between accuracy and robustness of these methods, and identify points of failures for future methods to improve upon.

%% file: sections/2_related_work.tex
\section{Related Work}
\begin{figure*}[t]
    \centering
   \includegraphics[width=\linewidth]{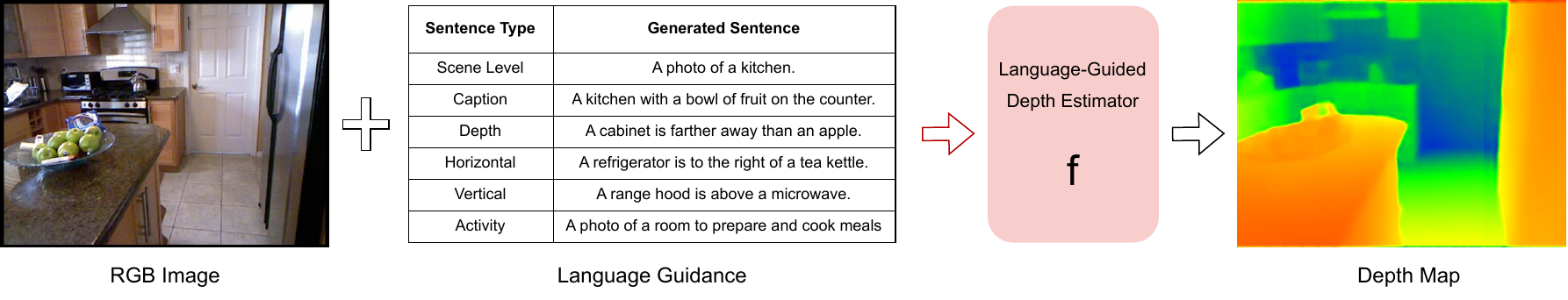}
    \caption{We systematically create additional knowledge for the depth estimator by leveraging intrinsic and low-level image properties. For each image we derive scene addendums, object and spatial level sentences along with semantic, activity based descriptions, and in supervised and zero-shot settings, quantify the effect of these sentences on monocular depth estimation. 
}%
    \label{fig:method_language}
\end{figure*}

\noindent\textbf{Monocular Depth Estimation.} 
Monocular depth estimation has a long history of methods ranging from hand-crafted feature extraction to deep learning techniques and, recently, language-guided estimation. 
\citet{saxena2005learning} designed depth estimators as Markov Random Fields with hierarchical image features, 
\citet{eigen2014depth} pioneered convolution networks for this task, and \citet{laina2016deeper} introduced fully convolutional residual networks for depth estimation.
\cite{qi2018geonet, xu2018pad, zhang2018joint} aim to jointly learn pixel-level dense prediction tasks such as depth estimation, semantic segmentation, and surface normal estimation in a multi-task learning setting.
Unsupervised methods \cite{garg2016unsupervised, godard2017unsupervised} have also been employed, using stereo images and multi-view perspective to guide learning.
More recently, VPD \cite{zhao2023unleashing}, TADP \cite{tadp}, EVP \cite{EVP} and DepthGen \cite{saxena2023monocular} have developed depth estimators that use diffusion-based models. Some standard datasets for monocular depth estimation include NYUv2 \cite{Silberman:ECCV12}, KITTI \cite{geiger2013vision}, Sun RGB-D \cite{song2015sun} and Cityscapes \cite{Cordts2016Cityscapes}, which vary based on their settings (indoor/outdoor), annotation type (sparse/dense) and sensor types.

\smallskip\noindent\textbf{Language Guidance for Lower Level Vision Tasks.}
Excellent results have been attained by CLIP \cite{radford2021learning} and ALIGN \cite{jia2021scaling} for zero-shot image recognition and OWL-ViT \cite{minderer2022simple} for open-vocabulary object detection.
GroupVIT and OpenSeg \cite{ghiasi2022scaling} perform open vocabulary semantic segmentation, supervised with high-level language, whereas ODISE \cite{xu2023open} performs open-vocabulary panoptic segmentation. OWL-ViT \cite{minderer2022simple} performs open-vocabulary object detection by aligning their pretrained, modality specific encoders with lightweight object-detectors and localization heads. 

\smallskip\noindent\textbf{Geometric Guidance for High-Level Vision Tasks.}
Low-level visual signals have been incorporated to solve high-level vision tasks such as visual question answering (VQA), image captioning, and image generation. \cite{banerjee2021weakly} make use of depth maps as additional supervision for VQA.
For 3D captioning, Scan2Cap \cite{chen2021scan2cap} proposes a message-passing module via a relational graph, where nodes are objects with edges capturing pairwise spatial relationship, while SpaCap3D \cite{wang2022spatiality} constructs an object-centric local 3D coordinate plan, incorporating spatial overlaps and relationships into its supervision. ControlNet \cite{zhang2023adding} conditions text-to-image generation based on low-level feedback from edge maps, segmentation maps, and keypoints. Florence \cite{florence} and Prismer \cite{prismer} leverage depth representations and semantic maps to develop general purpose vision systems.

\smallskip\noindent\textbf{Robustness Evaluation of Depth Estimation Models.} 
\citet{ranftl2020towards} study the robustness of depth estimation methods, by developing methods for cross-mixing of diverse datasets and tackle challenges related to scale and shift ambiguity across benchmarks. 
\cite{zhang2020adversarial, cheng2022physical} demonstrate that monocular depth estimation models are susceptible to global and targeted adversarial attacks with severe performance implications.
Our work studies the robustness of language-guided depth estimation methods.

%% file: sections/3_language_guided_depth_estimation.tex
\section{Language-Guided Depth Estimation}
\label{sec:prelims}

The use of natural language descriptions to facilitate low-level tasks is a new research direction.
Although at a nascent stage, early evidence from depth estimation suggests that language can indeed improve the accuracy of depth estimators.
This evidence comes from two recent approaches: VPD (visual perception with a pre-trained diffusion model) and TADP (text-alignment for diffusion-based perception) that show state-of-the-art results on standard depth estimation datasets such as NYU-v2 \cite{Silberman:ECCV12}. Our experiments are based on VPD thanks to open-source code. 

\subsection{Preliminaries}
The VPD model $f$ takes as input an RGB image $I$ and its scene-level natural language description $S$, and is trained to generate depth map $D_I$ of the input image: 
$D_I = f(I,S)$.

VPD has an encoder-decoder architecture. 
The encoding block consists of:
\begin{enumerate}[label=(\alph*),leftmargin=*]
    \item a frozen CLIP text encoder which generates text features of $S$, which are further refined by an MLP based Text Adapter \cite{gao2021clipadapter} for better alignment, and
    \item a frozen VQGAN \cite{esser2021taming} encoder which generates features of $I$ in its latent space.
\end{enumerate}
The cross-modal alignment is learnt in the U-Net of the Stable Diffusion model, which generates hierarchical feature maps. 
Finally, the prediction head, implemented as a Semantic FPN \cite{kirillov2019panoptic}, is fed these feature maps for downstream depth prediction, optimizing the Scale-Invariant Loss.

\paragraph{Format of Language Guidance:} Language guidance is provided via sentence $S$, which is a high-level description such as \textit{"an {\tt[ADJECTIVE]} of a {\tt[CLASS]}"}, where \texttt{[ADJECTIVE]} could be (photo, sketch, rendering) and \texttt{[CLASS]} is one of the 27 scene labels (bedroom, bathroom, office etc.) that each of the images in NYUv2 belong to. For each scene type, variations of templated descriptions are generated, encoded via CLIP, and averaged to generate its embedding. Finally, each image $I$ based on its scene type, is mapped to the generated embedding, which is considered to be high-level knowledge about $I$.

\subsection{Diverse Sentence Creation}
While scene-level description have been successfully used for low-level tasks, 
the effect of different language types remains under-explored. 
In this subsection, we outline our framework of creating a diverse set of sentence types in order to evaluate and better understand the influence of language in improving depth estimation.
We define $S$ to be the baseline scene-level description (as used in VPD) of image $I$.
Figure \ref{fig:method_language} displays our workflow of sentence generation.

\vspace{-0.1cm}
\subsubsection{Sentences Describing Spatial Relationships} 
For a given image, our goal is to generate sentences that represent object-centric, low-level relationships in that image. 
Humans approximate depth through pictorial cues which includes relative relationships between objects. 
We focus on generating pairwise relationships between objects without creating complex sentences that would necessitate the model to engage in additional, fine-grained object grounding. 
These descriptions, which mirror human language patterns, explicitly contain depth information and could be potentially beneficial for improving depth estimation.

Specifically, for all images $I$, we  have semantic and depth ground-truth annotations at an instance and object-level across the dataset. Given this information, we generate sentences that describe the spatial relationship between a pair of objects, in an image. 
We consider 3D relationships, i.e. depth-wise, horizontal and vertical relationships between an object pair, and thus the set of all spatial relationships $R$ is defined as $\{front, behind, above, below, left, right\}$.  
Given $I$, and two objects $A$ and $B$ present in it, twelve relationships can be generated between them as given below:

\newcommand{\relationships}{%
   \textit{\textcolor{black!60} front(A,B), behind(A,B), front(B,A), behind(B,A), above(A,B), below(A,B), above(B,A), below(B,A), right(A,B), left(A,B), right(B,A), left(B,A)}%
}
\begin{tcolorbox}[colback=white, colframe=black, rounded corners, width=\linewidth, boxrule=0.1mm]
  \parbox{\linewidth}{\centering \small \relationships}
\end{tcolorbox}

\noindent\textbf{Relationship Extraction:}
For an object $A$, let $(X_a, Y_a)$ be the coordinates of its centroid, $R_a$ be it's maximum radius, $(\mu_a, \sigma_a, M_a)$ be the mean, standard deviation, and maximum object depth. 
Between A and B, a horizontal relationship is created if : $ |(Y_{a}-Y_{b}) > \lambda  \times (R_{a}+R_{b})| $, 
where 
$\lambda$ controls the amount of overlap allowed between the two objects. Thereafter, A is to the left of B if $ (Y_{a} < Y_{b}) $, and otherwise. 
Similarly, a vertical relationship is created if : $ |(X_{a}-X_{b}) > \lambda  \times (R_{a}+R_{b})| $ and A is above B if $ (X_{a} < X_{b}) $. 
Finally, a depth relationship is created if : $ |(\mu_{a}-\mu_{b}) > ((M_{a} - \mu_{a}) + (M_{b} - \mu_{b}))| $. Thus, A is closer than B if $ (\mu_{a} + \sigma_{a} < \mu_{b} + \sigma_{b}) $, else A is farther than B.
Having found $R$, we map the relationships into one of the templated sentences, as shown below:
\newcommand{\sentences}{%
   \textit{\textcolor{black!60} A is in front of B, A is closer than B,
A is nearer than B, A is behind B, A is farther away than B, A is more distant
than B, A is above B, A is below B, A is to the right of B , A is to the left
of B}%
}
\begin{tcolorbox}[colback=white, colframe=black, rounded corners, width=\linewidth, boxrule=0.1mm]
  \parbox{\linewidth}{\centering \small \sentences}
\end{tcolorbox}
Similar templates have also been recently leveraged in evaluation of T2I Models \cite{visor, t2i_compbench} and for language grounding of 3D spatial relationships \cite{goyal2020rel3d}.

Thus, given an image $I$, we test the model’s performance by generating sentences that explicitly encode depth information and relative spatial locations of objects in the image. 
Intuitively, depth estimation could benefit from information about objects and their spatial relationships as compared to the scene-level descriptions.
\subsubsection{Image Captions and Activity Descriptions}
\paragraph{Image captions.}
We generate captions corresponding to each image, which can be characterized as providing information in addition to scene level description. The rationale is to evaluate the performance of the model, when furnished a holistic scene level interpretation, more verbose in comparison to the baseline sentence $S$. Captions may still capture object-level information but will not furnish enough low-level information for exact localization. 

\smallskip\noindent\textbf{Activity descriptions.}
To test the semantic understanding of a scene, we modify the scene name in $S$, replacing it with a commonplace activity level description of
a scene. For example, "A picture of a "\textit{kitchen}" is replaced with "A picture of a "\textit{room to prepare and cook meals}". Full list of transformations are presented in the Appendix. We curate these descriptions via ChatGPT.

To summarize, for an image $I$, we now possess sentences that delineate various aspects of it, encompassing object-specific, scene-specific, and semantic details. These {\tt<image, text>} pairs are used in the next sections to quantify the impact of language.

%% file: sections/4_experiments.tex
\section{Measuring the Effect of Language Guidance} 
\label{sec:language}

In the following subsections, we quantify the importance of language conditioning in supervised and zero-shot settings. Following standard metrics, we report results on the Root Mean Square Error (RMSE), Absolute Mean Relative Error (Abs. REL), Absolute Error in log-scale (Log$_{10}$), and the percentage of inlier pixels $\delta^i$ with a threshold of 1.25$^i$ (i=1,2,3).  We use the flip and sliding window techniques during testing. For all subsequent tables, \textbf{Bold} and \underline{underlined} values indicate best and second-best performance, respectively. All our sentences are generated on the 1449 (Train = 795, Test = 654) images from the NYUv2 official dataset, which contain dense annotations. The maximum depth is set to 10m and  set $\lambda$ = 1.

\input{tables/supervised_comparison}
\begin{figure}
    \centering 
    \includegraphics[width=\linewidth]{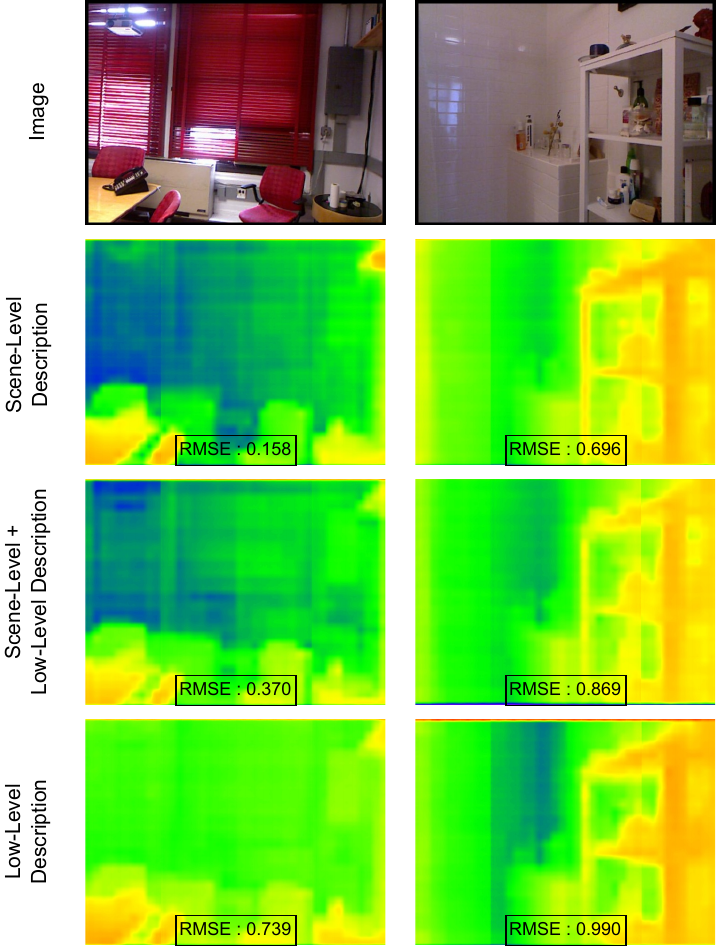}
    \caption{Comparison of depth maps across the three models trained under the supervised setting as described in Table \ref{tab:supervised_Setting}. Low-level sentences induce hallucinations in the model; leading to large errors 
    and false positive long-range depth estimates 
    }
    \label{fig:supervised}
\end{figure}

\subsection{Supervised Experiments} 
    In this setting, we answer, \textbf{does training on low-level language help?} 
    We find that when trained and evaluated with additional low-level language, model performance decreases (Table \ref{tab:supervised_Setting}). 
    Apart from the baseline model, we train two more models s.t. for each $I$
    \begin{enumerate}[label=(\alph*)]
        \item baseline sentence $S$ and 1-3 supplementary sentences containing low-level relationships are used, and
        \item 4-6 sentences where only spatial relationships are used.
    \end{enumerate} 
    Compared to only low-level sentences, combining low-level with scene-level sentences deteriorates performance.
    This indicates that current approaches interpret language only when it is coarse-grained and require scene-level semantics for optimal performance. 
    We present examples in Figure \ref{fig:supervised}.

\input{tables/main_table}

\begin{figure*}
    \centering 
    \includegraphics[width=\linewidth]{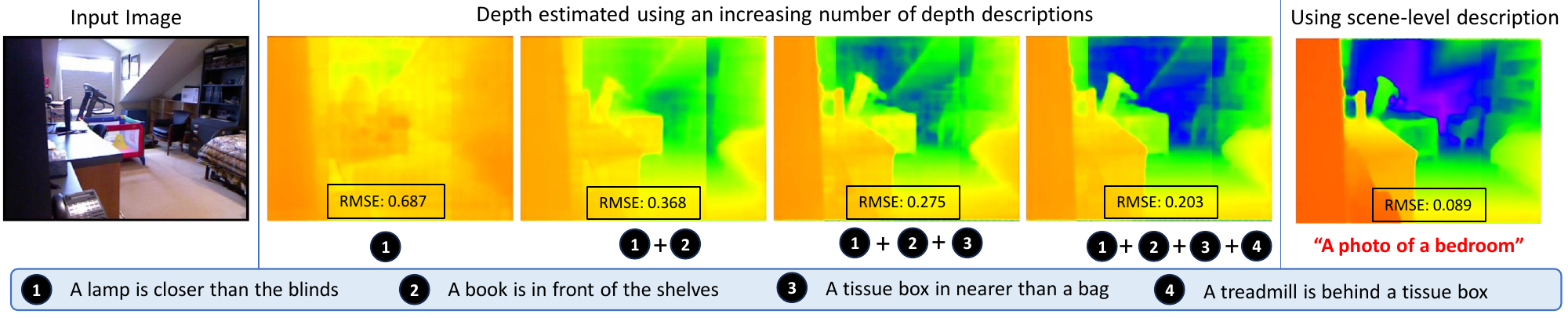}
    \caption{From \textbf{left} to \textbf{right}, as more bottom-up scene-level information is provided, the model's depth predictions move closer to the baseline predictions made with scene-level sentences.
    The plot below shows performance improvement across all metrics.}%
    \label{fig:image_2}
\end{figure*}

\begin{figure}
    \centering 
    \includegraphics[width=\linewidth]{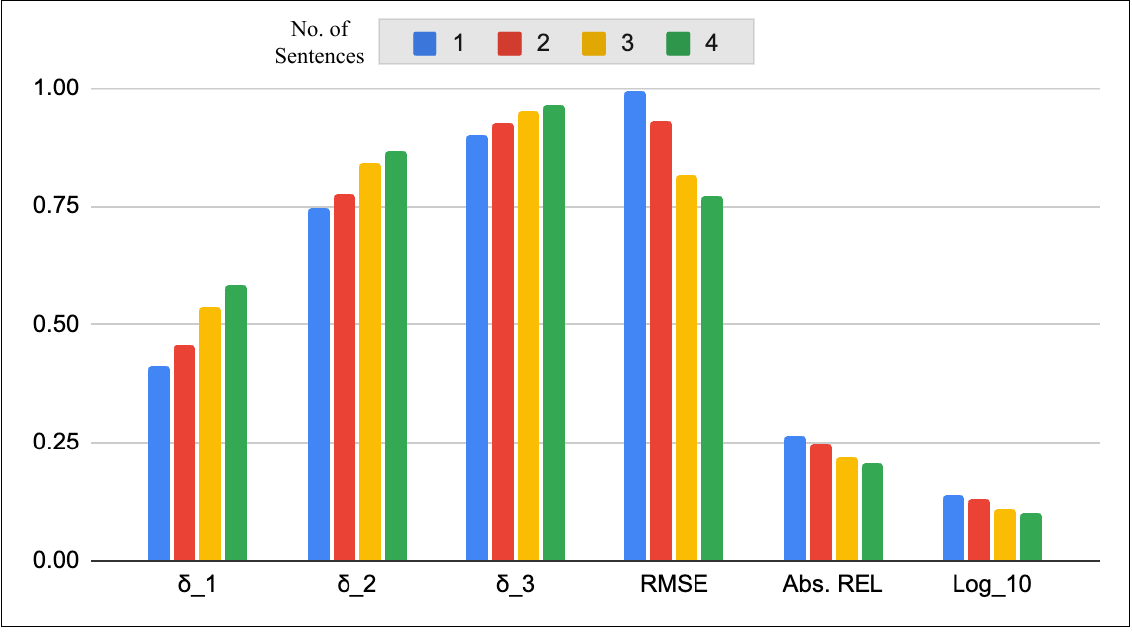}
    \caption{
    As more depth descriptions  are provided, performance improves, signifying scene-level alignment.}%
    \label{fig:depth_ablation}
\end{figure}

\subsection{Zero-Shot Findings}  
All zero-shot experiments are performed on the open-source VPD model.
Language embeddings are generated via CLIP  with an embedding dimension of 768, and image captions are generated using the BLIP-2-OPT-2.7b model \cite{li2022blip}.

\vspace{-5pt}
\paragraph{Impact of Sentence Types:} We evaluate VPD on our created sentences as shown in Table \ref{tab:main_table}. Sentences are generated for 518 images from the NYUv2 test split, considering only images where at least 1 depth, vertical, and horizontal sentence can be extracted. To avoid ambiguity, we only consider sentences between unique objects in a scene.  As mentioned in Section \ref{sec:prelims}, the original method averages out multiple scene-level descriptions, which are created using 80 ImageNet templates \cite{xu2022groupvit}, and leverages the mean CLIP embedding as high level information. Following the original method, $^\ast$ in Table \ref{tab:main_table} represents the set-up, where for every $I$, we generate embeddings by stacking the mean baseline embedding and our sentence embeddings while in $^\oplus$, for every sentence $T$ $\in$ the ImageNet Template, we concatenate $T$ and our sentences, and compute its CLIP embedding. The key differentiator is that in the former, the weight of the baseline (scene-level) description and the other sentences are equal, while in the latter, the low-level sentences have more prominence by virtue of them being added for each sentence in the template.

We re-affirm our initial findings through Table \ref{tab:main_table}. the method maintains its optimal performance only when presented with scene-level sentences. Counter-intuitively, the performance gradually worsens as additional knowledge (both high and low-level) is provided. Even other forms of high-level language seem to deteriorate performance. Next, we observe a clear bias towards scene level description. For example, (Baseline + Caption) and (Caption Only) always outperform (Baseline + Caption + X) and (Depth/2D Only). This claim can be further underlined by the $\Delta$ decrease in performance from $^\ast$ to $^\oplus$, showing a distinct proclivity towards scene-level descriptions.

In Figure \ref{fig:image_1}, we present a visual illustration of the scene bias that persists in these methods. The least performance drop is only seen in cases where additional high-level knowledge is provided. Iterative addition of low-level knowledge adversely affects the performance model's spatial comprehension. The model appears to completely apply a smooth depth mask in certain circumstances (such as vertical only), completely disregarding image semantics. 
\vspace{-5pt}
\paragraph{Does Number of Sentences Matter?} 
We find that using multiple low-level sentences, each describing spatial relationships, helps performance -- performance is correlated with number of such sentences used.

This can be attributed to more sentences offering better scene understanding. 
We find \textit{again}, that model needs enough "scene-level" representation to predict a reasonable depth map as observed in Figure \ref{fig:depth_ablation}. 
When the number of sentences is increased from 1 to 4 we observe a 41\% increase and a 30\% decrease in $\delta^1$ and RMSE, respectively. We present an illustrative example in Figure \ref{fig:image_2}, and observe that as the number of sentences are increased, the depth prediction iteratively aligns with the predictions from the scene-level sentences.

\vspace{-5pt}
\paragraph{Understanding of Semantics.} When we use sentences at the activity level and compare it against the baseline scene-level sentences (Table \ref{tab:activity}), we see that the RMSE increases by 17\%. Despite our transformations being extremely simple, we find mis-alignment between the semantic and scene space. A method exclusively tuned towards scene-level alignment, lacking semantic understanding would lead to unwanted failures, which would be particularly detrimental in a real-world setting. 

\input{tables/activity}

\input{tables/clip_score}

\subsection{Potential Explanations for Failure Modes}
Language-guided depth estimation methods that we discussed above have \textbf{2} major components, a trainable Stable Diffusion (U-Net) Encoder and a frozen CLIP Encoder. We take a detailed look into each of them to find answers that explain these shortcomings.

The lack of understanding of spatial relationships of Diffusion-based T2I models is well studied by VISOR \cite{visor} and T2I-CompBench \cite{t2i_compbench}. Studies \cite{chatterjee2024spade} show that the cross-attention layers of Stable Diffusion lack spatial faithfulness to the input prompt; these layers itself are  used by VPD to generate feature maps which could explain the current gap in performance. Similarly, to quantify CLIP's understanding of low-level sentences, we perform an experiment where we generate the CLIPScore \cite{clipscore} between RGB Images from NYUv2 and our generated ground-truth sentences. We compare the above score, by creating adversarial sentences where we either switch the relationship type or the object order, keeping the other fixed. We find (Table \ref{tab:clip_score}) that \textbf{a)} CLIPScore for all the combinations are low but more importantly, \textbf{b}) the $\Delta$ difference between them is negligible; with the incorrect sentences sometimes yielding a higher score. (highlighted in \textcolor{red}{red}).
Similar findings have been recently reported for VQA and image-text matching by Kamath et al. \cite{kamath2023s} and Hsu et al. \cite{hsu}. 

To summarize, while it is tempting to use language guidance in light of the success of LLMs, we show that current methods are prone to multiple modes of failures. 
In order to facilitate practical deployments, these techniques must be resilient and function well in novel environments and domain shifts, which we study next.

%% file: tables/supervised_comparison.tex
\begin{table}[t]
\centering
\LARGE
\resizebox{\linewidth}{!}{
\begin{tabular}{@{}lcccccc@{}}
\toprule
\textbf{Sentence Type} &  $\delta_{1}$ ($\uparrow$)  &  $\delta_{2}$ ($\uparrow$)  &  $\delta_{3}$ ($\uparrow$) &  RMSE ($\downarrow$) &  Abs.\ REL  ($\downarrow$) &  Log$_{10}$ ($\downarrow$) \\
\midrule
 Scene-Level (Baseline) &  \textbf{0.861} & \textbf{0.977} & \textbf{0.997}  & \textbf{0.382} & \textbf{0.122} & \textbf{0.050} \\
Scene-Level + Low-Level & 0.819 & 0.964 & 0.993 & 0.440 & 0.149 & 0.059 \\
Only Low-Level &  \underline{0.844} & \underline{0.969} & \underline{0.994} & \underline{0.424} & \underline{0.135} & \underline{0.055} \\
\bottomrule
\end{tabular}
}
\caption{
Counter-intuitively, training with spatial sentences impairs performance compared to training with scene-level descriptions, limiting the efficacy of language-guided depth estimation.
}
\label{tab:supervised_Setting}
\end{table}

%% file: tables/main_table.tex
\begin{table*}[t]
\small
\centering
\begin{tabular}{lcccccc}
\toprule
\textbf{Sentence Type} & $\delta_{1} (\uparrow)$ & $\delta_{2} (\uparrow)$ & $\delta_{3} (\uparrow)$ & RMSE ($\downarrow$) & Abs. REL ($\downarrow$) & $Log_{10}$ ($\downarrow$) \\
\midrule
Scene-Level & \textbf{0.962} & \textbf{0.994} & \textbf{0.999} & \textbf{0.252} & \textbf{0.068} & \textbf{0.029} \\
Scene-Level  + Caption $^\ast$ & \underline{0.950} & \underline{0.993} & \underline{0.998} & \underline{0.279} & \underline{0.076} & \underline{0.033} \\
Scene-Level + Caption + Depth $^\ast$ & 0.932 & 0.992 & 0.998 & 0.311 & 0.084 & 0.037 \\
Scene-Level  + Caption + Depth + 2D $^\ast$  & 0.864 & 0.973 & 0.993 & 0.403 & 0.109 & 0.050 \\
Scene-Level  + Caption $^\oplus$ & 0.916 & 0.986 & 0.997 & 0.347 & 0.092 & 0.041 \\
Scene-Level  + Caption + Depth  $^\oplus$ & 0.878 & 0.980 & 0.994 & 0.399 & 0.105 & 0.048 \\
Scene-Level  + Caption + Depth + 2D $^\oplus$ & 0.849 & 0.973 & 0.994 & 0.443 & 0.115 & 0.053 \\
Caption Only & 0.827 & 0.961 & 0.988 & 0.474 & 0.127 & 0.059 \\
Depth Only & 0.372 & 0.696 & 0.878 & 1.045 & 0.284 & 0.153 \\
Vertical Only & 0.260 & 0.583 & 0.824 & 1.223 & 0.329 & 0.185 \\
Horizontal Only & 0.332 & 0.633 & 0.838 & 1.148 & 0.306 & 0.170 \\
\bottomrule
\end{tabular}
\caption{In a zero-shot setting, VPD's performance is highest with baseline scene-level sentences. However, performance drops when more detailed, low-level information is introduced, as indicated by an increase in RMSE.}
\label{tab:main_table}
\end{table*}

%% file: tables/activity.tex
\begin{table}[t]
\centering
\LARGE
\resizebox{\linewidth}{!}{
\begin{tabular}{@{}lcccccc@{}}
\toprule
\textbf{Setup} &  $\delta_{1}$ ($\uparrow$)  &  $\delta_{2}$ ($\uparrow$)  &  $\delta_{3}$ ($\uparrow$) &  RMSE ($\downarrow$) &  Abs. REL  ($\downarrow$) &  Log$_{10}$ ($\downarrow$) \\
\midrule
Scene Level & \textbf{0.963} & \textbf{0.994 }& \textbf{0.998} & \textbf{0.254} & \textbf{0.069} & \textbf{0.029} \\
Activity Level & \underline{0.936}	& \underline{0.991} &	\underline{0.998}	& \underline{0.297} &	\underline{0.085}	& \underline{0.036} \\
\bottomrule
\end{tabular}
}
\caption{In contrast to scene level sentences, semantics denoting activity level sentences result in a performance decline.}
\label{tab:activity}
\end{table}

%% file: tables/clip_score.tex
\begin{table}
\centering
\large
\resizebox{\linewidth}{!}{
\begin{tabular}{@{}lccccc@{}}
\toprule
\textbf{Relationship} &  \stackanchor{Original}{Sentence} & \stackanchor{Relationship}{Switch} & \stackanchor{Object}{Switch} & $\Delta_{orig. - rel.}$ & $\Delta_{orig. - obj.}$ \\
\midrule
Horizontal & 25.675 & 25.665 & 25.699 & 0.009 &  \color{red}-0.024 \\
Vertical & 23.138 & 23.161 & 23.206 & \color{red}-0.023 &  \color{red}-0.068 \\
Depth & 23.613 & 23.562 & 23.537 & 0.050 & 0.075 \\
\bottomrule
\end{tabular}
}
\caption{CLIP struggles at differentiating between various spatial sentences, often producing higher scores for incorrect sentences spatial relationships.}
\label{tab:clip_score}
\end{table}

%% file: sections/5_robustness.tex
\section{Robustness and Distribution Shift} \label{sec:robust}
To assess the impact of the language signal under adversarial conditions, we setup the following experiments where we compare vision-only methods with VPD : 

\begin{figure*}[t]
    \centering
   \includegraphics[width=\linewidth]{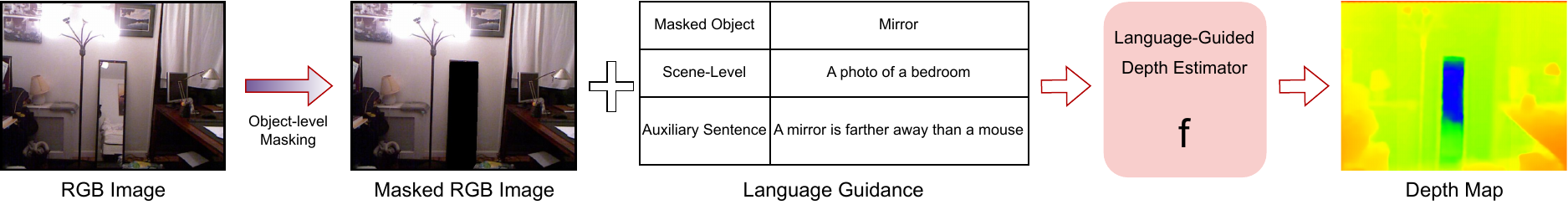}
    \caption{We mask an unique object in the image, using ground-truth segmentation annotations. To compensate for the loss of image signal, we provide additional knowledge about the masked object, followed by performing depth estimation.
}%
\label{fig:method_masking}
\end{figure*}

\medskip\noindent\textbf{Masking :} As shown in Figure \ref{fig:method_masking}, we perturb the image $I$ in this setup, by masking an object in the image space. To offset the image-level signal loss, we include a language-level input specifying the precise relative position of the masked object with another object. We find that \textbf{vision-only models are more resilient to masking} in comparison to language-guided depth estimators. We compare AdaBins and VPD (Table \ref{tab:mask}) and find that the latter's $\Delta$ drop in performance is significantly more in comparison to its baseline performance. Despite leveraging additional information about the relative spatial location, VPD is less resilient in comparison to AdaBins. Following previous trends, we also find that the performance deteriorates significantly when scene-level information is removed. 
\input{tables/mask}

\medskip
In the following experiments, we compare VPD with AdaBins \cite{Farooq_Bhat_2021}, MIM-Depth \cite{xie2022revealing} and IDisc \cite{piccinelli2023idisc}.
\vspace{-5pt}
\paragraph{Scene Distribution Shift under the Supervised Setting:} We define a new split of the NYUv2 dataset, where the train and test set have 20 and 7 non-overlapping scenes, with a total of 17k and 6k training and testing images. 
With this configuration, we train all the corresponding models and benchmark their results and adhere to all of the methods’ original training hyper-parameters, only slightly reducing the batch size of IDisc to 12. 

Although VPD follows MIM-Depth as the 2nd-best performing model, we find that VPD has the largest performance drop amongst its counterparts, \textbf{107\%}, when compared to their original RMSE (Table \ref{tab:supervised_ood}) . Since training is involved, we also allude to the \# of trainable parameters to quantify the trade-off between performance and efficiency of the respective models.

\input{tables/label_shift_test}

\medskip\noindent\textbf{Zero-shot Generalization across Datasets:} 
We perform zero-shot experiments on the models trained on NYUv2 and test its performance on Sun RGB-D \cite{song2015sun}, without any further fine-tuning. 
Sun RGB-D contains 5050 testing images across 42 different scene types. 
We create sentences of the form \textit{``a picture of a {\tt[SCENE]}''}.

We find that \textbf{language guided depth estimation methods struggle to generalize across image datasets}. VPD has a 20\% higher RMSE in comparison to MIM-Depth, the best performing model (Table \ref{tab:domain_indoor_shift}). This occurs even though Sun RGB-D and NYUv2 are both indoor datasets with a 50\% overlap of scene type similarity.

This difference in performance between the two categories of models likely occurs because in language guided depth estimators, the model is forced to learn correlations between an in-domain and its \textit{high-level} description. It cannot, therefore, map its learned representation to new data when an out-of-domain image with an unseen description is presented. On the contrary, vision-only depth estimators are not bound by any \textit{language} constraints, and hence learn a distribution which better maps images to depth.

\begin{figure}[t]
    \centering 
    \includegraphics[width=\linewidth]{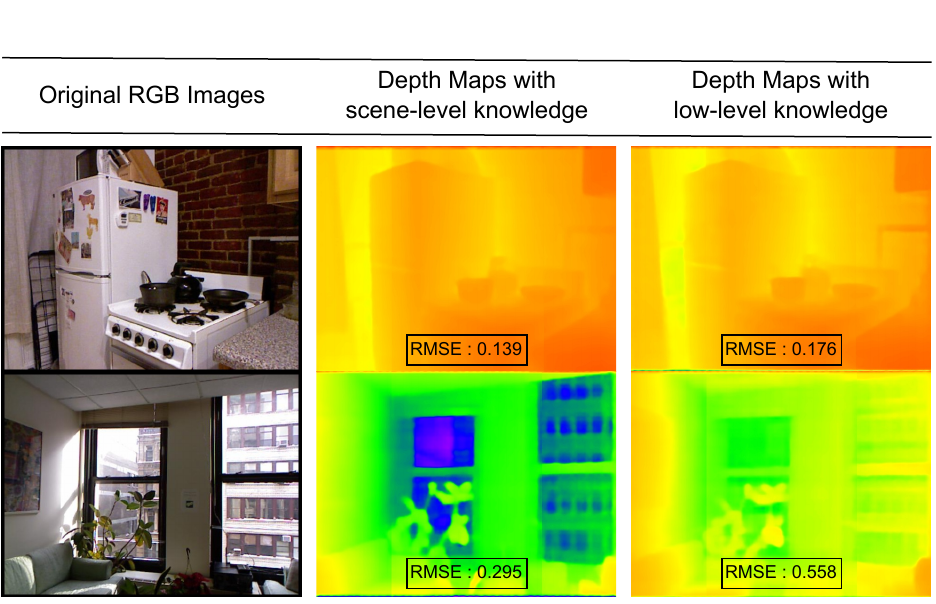}
    \caption{
    When provided with low-level knowledge such as, "a plant is to the right of a picture" about a scene with a large variation in depth values, VPD does not perform well (\textbf{bottom}), as compared to when the scene is more localized (\textbf{top}).
    }
    \label{fig:image_3}
\end{figure}

We also identify interesting correlations between the impact of low-level knowledge and the intrinsic properties of an image. As shown in Figure \ref{fig:image_3}, the drop in performance of VPD is substantially high where the original RGB image has a large variation in depth.

%% file: tables/mask.tex
\begin{table}[t]
\centering
\large
\resizebox{\linewidth}{!}{
\begin{tabular}{@{}llccc@{}}
\toprule
\textbf{Model, Image} & \textbf{Sentence} & $\Delta$ $\delta_{1}$ ($\downarrow$) &$\Delta$ RMSE ($\downarrow$) & $\Delta$ Abs. REL  ($\downarrow$) \\
\midrule
VPD & Scene-Level + Depth & \underline{0.062} & \underline{0.093} & \underline{0.024}  \\ 
VPD & Depth & 0.586 & 0.794 & 0.213 \\
AdaBins & N/A & \textbf{0.008} & \textbf{0.007} & \textbf{0.002}  \\
\bottomrule
\end{tabular}
}
\caption{Under the masked image setting, we compare $\Delta$ decrease of VPD with AdaBins (vision-only depth estimator). AdaBins is significantly more robust to masked objects than VPD.}
\label{tab:mask}
\end{table}

%% file: tables/label_shift_test.tex
\begin{table}[t]
\centering
\large
\resizebox{\linewidth}{!}{
\begin{tabular}{@{}lccccc@{}}
\toprule
\textbf{Method} & \stackanchor{Params}{(Millions)} & $\delta_{1}$ ($\uparrow$) & RMSE ($\downarrow$)& $\Delta_{RMSE(original)} \% (\downarrow)$ &Abs. REL ($\downarrow$) \\
\midrule
AdaBins & 78 & 0.763  & 0.730 &  100.54& 0.168 \\
MIM-Depth & 195 & \textbf{0.872} & \textbf{0.527} & \textbf{83.62}&\textbf{0.115}  \\
IDisc & 209 & 0.836 & 0.609 & \underline{94.56} &0.129 \\
VPD & 872 & \underline{0.867} & \underline{0.547} & 107.48 &\underline{0.121} \\
\bottomrule
\end{tabular}
}
\caption{Comparison of VPD and Vision-only models in the supervised, scene distribution setting. When evaluated on novel scenes, VPD has the largest drop in performance, compared to its baseline.}
\label{tab:supervised_ood}
\end{table}

\begin{table}[t]
\centering
\large
\resizebox{\linewidth}{!}{
\begin{tabular}{@{}lcccc@{}}
\toprule
\textbf{Method} & $\delta_{1}$ ($\uparrow$) & RMSE ($\downarrow$)& $\Delta_{RMSE(original)} \%$ ($\downarrow$) & Abs. REL ($\downarrow$)  \\
\midrule
AdaBins & 0.768 & 0.476 & 30.76&0.155 \\
MIM-Depth & \textbf{0.857} & \textbf{0.367} & \underline{27.87} &\underline{0.132} \\
IDisc & \underline{0.838} & \underline{0.387} & \textbf{23.64} &\textbf{0.128} \\
VPD & 0.786 & 0.442 & 74.01 &0.143 \\
\midrule
\end{tabular}
}
\caption{Comparative results between VPD and Vision-only models while zero-shot testing on the Sun RGB-D dataset.}
\vspace{-0.2cm}
\label{tab:domain_indoor_shift}

\end{table}

%% file: sections/6_conclusion.tex
\section{Conclusion}

Applying natural language priors to depth estimation opens new possibilities for bridging language and low-level vision. However, we find that current methods only work in a restricted setting with scene-level description, but do not perform well with low-level language, lack understanding of semantics, and possess a strong scene-level bias. Compared to vision-only models, current language-guided estimators are less resilient to directed adversarial attacks and show a steady decrease in performance with an increase in distribution shift. 
An examination of the causes of these failures reveals that foundational models are also ineffective in this context. As low-level systems are actively deployed in real-world settings, it is imperative to address these failures and to investigate the role of language \textit{in depth}. 
The findings from the paper could guide future work into better utilization of language in perception tasks.

%% file: sections/7_acknowledgement.tex
\section*{Acknowledgements}
The authors acknowledge Research Computing at Arizona State University for providing HPC resources and support for this work.
This work was supported by NSF RI grants \#1750082 and \#2132724. 
The views and opinions of the authors expressed herein do not necessarily state or reflect those of the funding agencies and employers. 

%% file: suppl.tex
\clearpage
\appendix
\maketitlesupplementary

\section{Activity-Level Transformations}
Table \ref{tab:activity_level_table} presents the transformations we leveraged to evaluate the semantic, activity level understanding of language-guided depth estimators, as mentioned in Section \ref{sec:language}. We curate these descriptions from ChatGPT.

\section{Train-Test Split : Out-of-Distribution Supervised Setting}
In Table \ref{tab:class_split_ood}, we present the new train-test split as described in Section \ref{sec:robust}, for the supervised setting. Out of 24,231 images, 17,841 were used in training , while the remaining 6,390 were used in testing.

\section{Scene-Level Sentence Type Distribution}
\begin{figure*}[t]
    \centering
    \includegraphics[width=\linewidth]{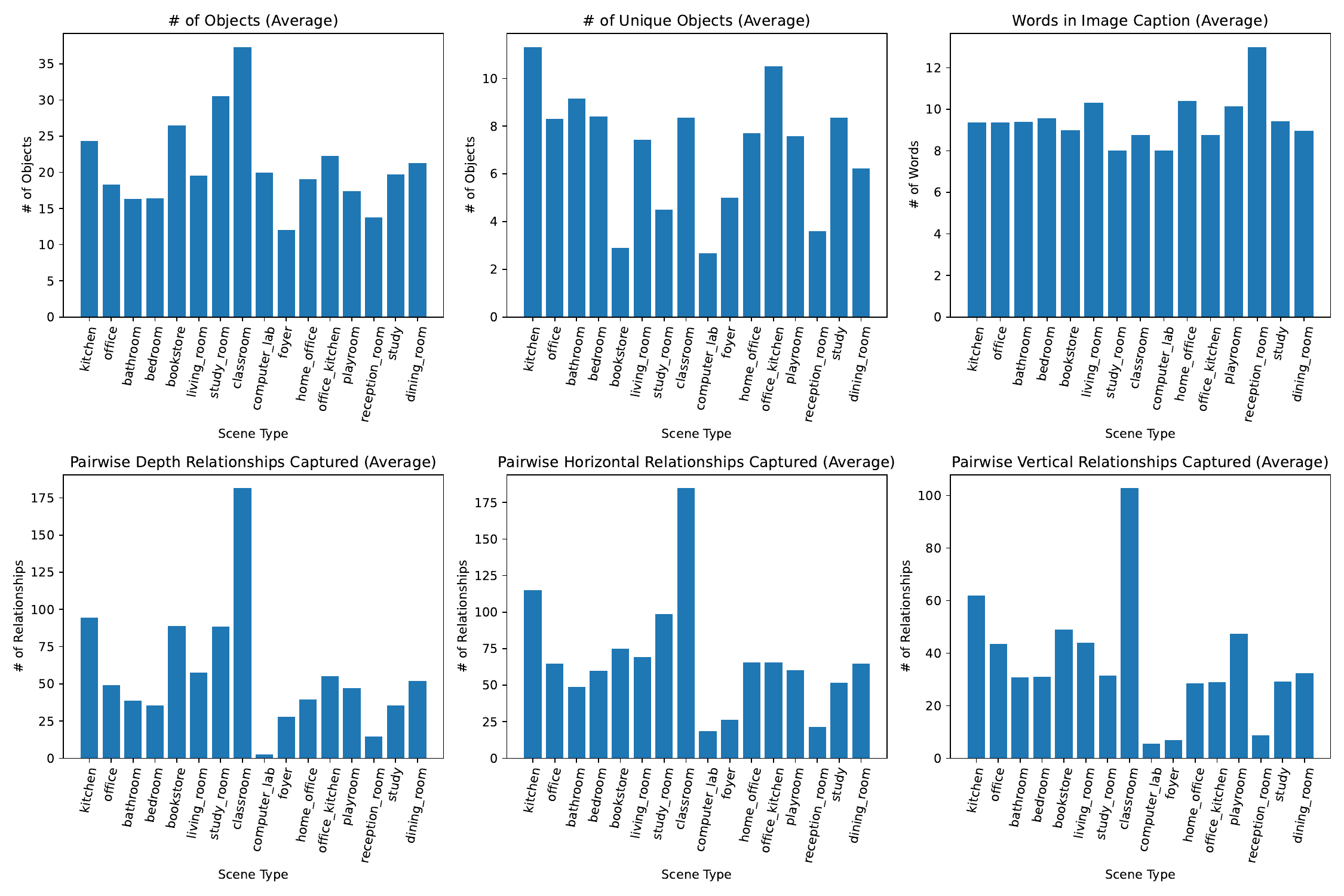}
    \caption{Graphical illustration of the average number of objects (complete and unique), average number of words in an image caption and the number of spatial relationships captured, across scene types. Results shown on NYUv2 test split.}%
    \label{fig:language_plots}
\end{figure*}

We present in Figure \ref{fig:language_plots}, an overall distribution of objects and the corresponding sentences created by our framework, across scenes on the NYUv2 Test Split. We use the NLTK tokenizer for calculating the average number of words in a caption. For the pairwise relationships, we present results on all objects in a scene, irrespective of their uniqueness. Hence, we notice a direct correlation between the \# of objects and the \# of relationships across scenes. Also, interestingly we find that the \# of vertical relationships are lesser in comparison to depth and horizontal; this can be attributed to NYUv2 being an indoor dataset having lesser variation in height.

\input{tables/room_activity_description}
\input{tables/ood_class_split}

\section{Distribution of Scenes in SUN RGB-D }
\begin{figure*}[]
    \centering
    \includegraphics[width=\linewidth]{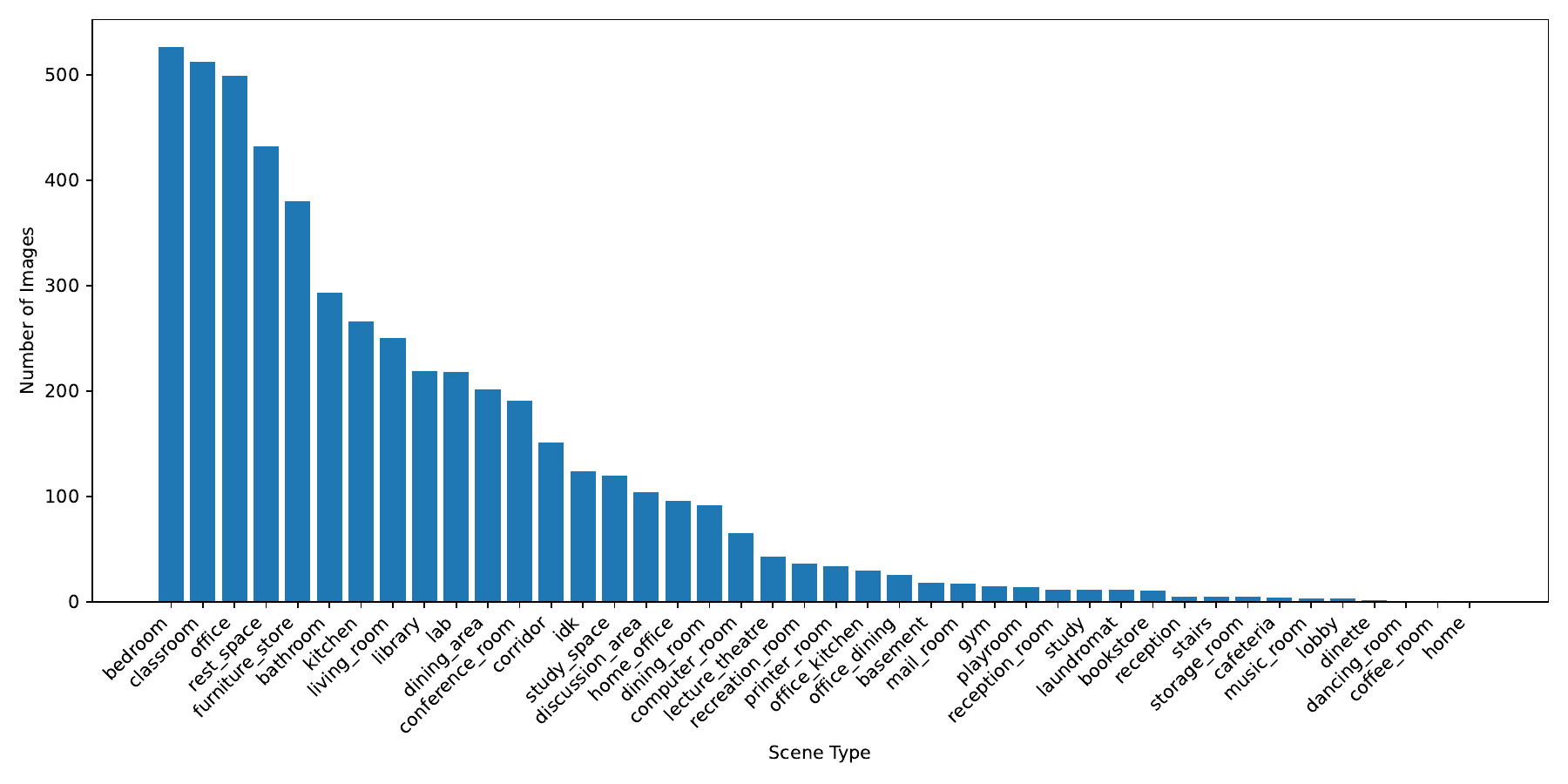}
    \caption{Number of instances per scene types in the SUN RGB-D dataset.}%
    \label{fig:sunrgbd}
\end{figure*}

In Figure \ref{fig:sunrgbd}, we illustrate the \# of images for a given scene type in the Sun-RGBD dataset. As shown in Section \ref{sec:robust}, despite having 50\% overlap in the scene-type between its training and testing distribution, VPD has the largest drop in performance.

\section{Additional Illustrations}
\begin{figure*}[]
    \centering
    \includegraphics[width=\linewidth]{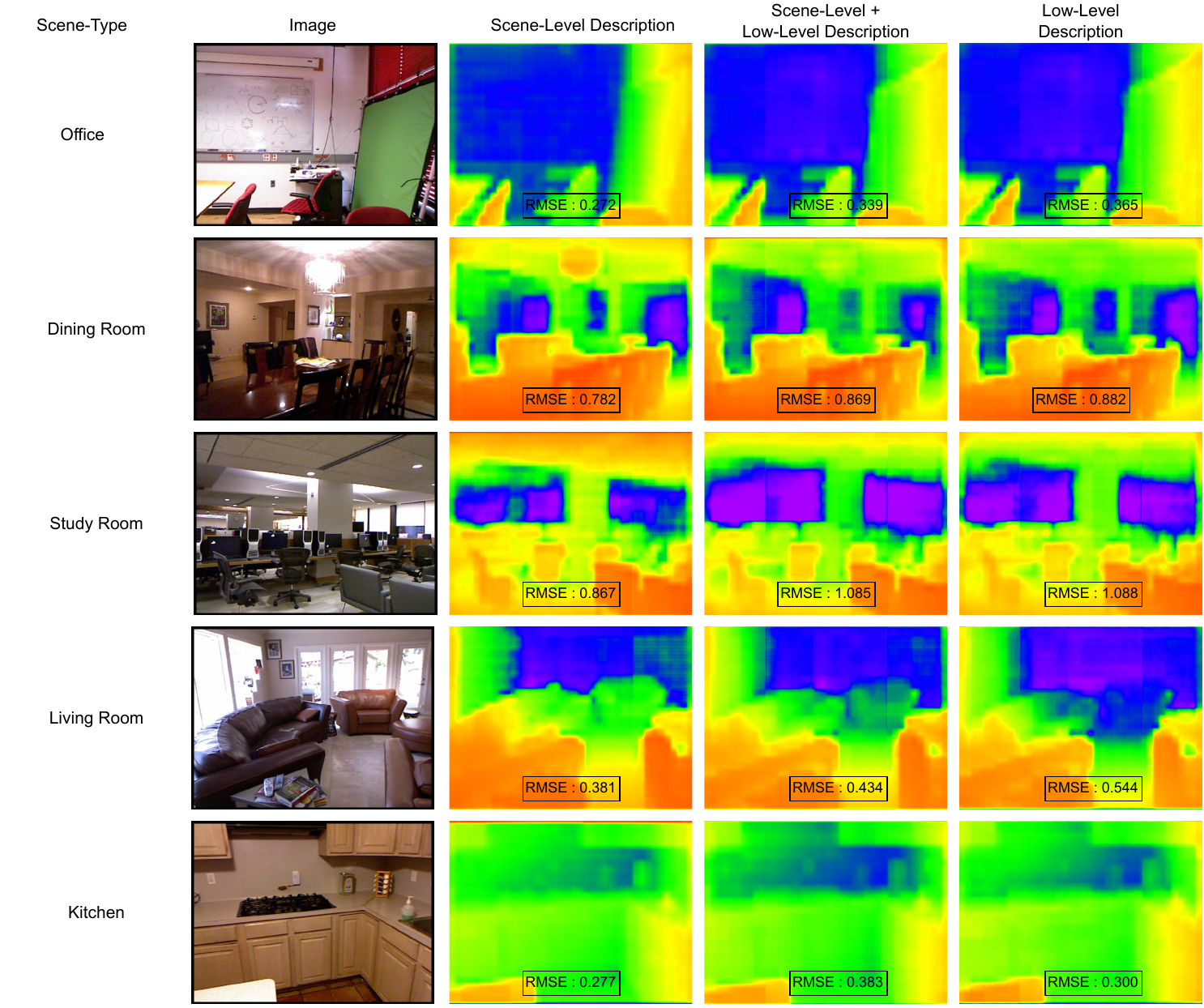}
    \caption{Comparison of Generated Depth Maps, trained in a \textbf{supervised} setting, across 5 different scene-types and 3 kinds of natural language guidance. }%
    \label{fig:supervised_supp}
\end{figure*}

\begin{figure*}[]
    \centering
    \includegraphics[width=\linewidth]{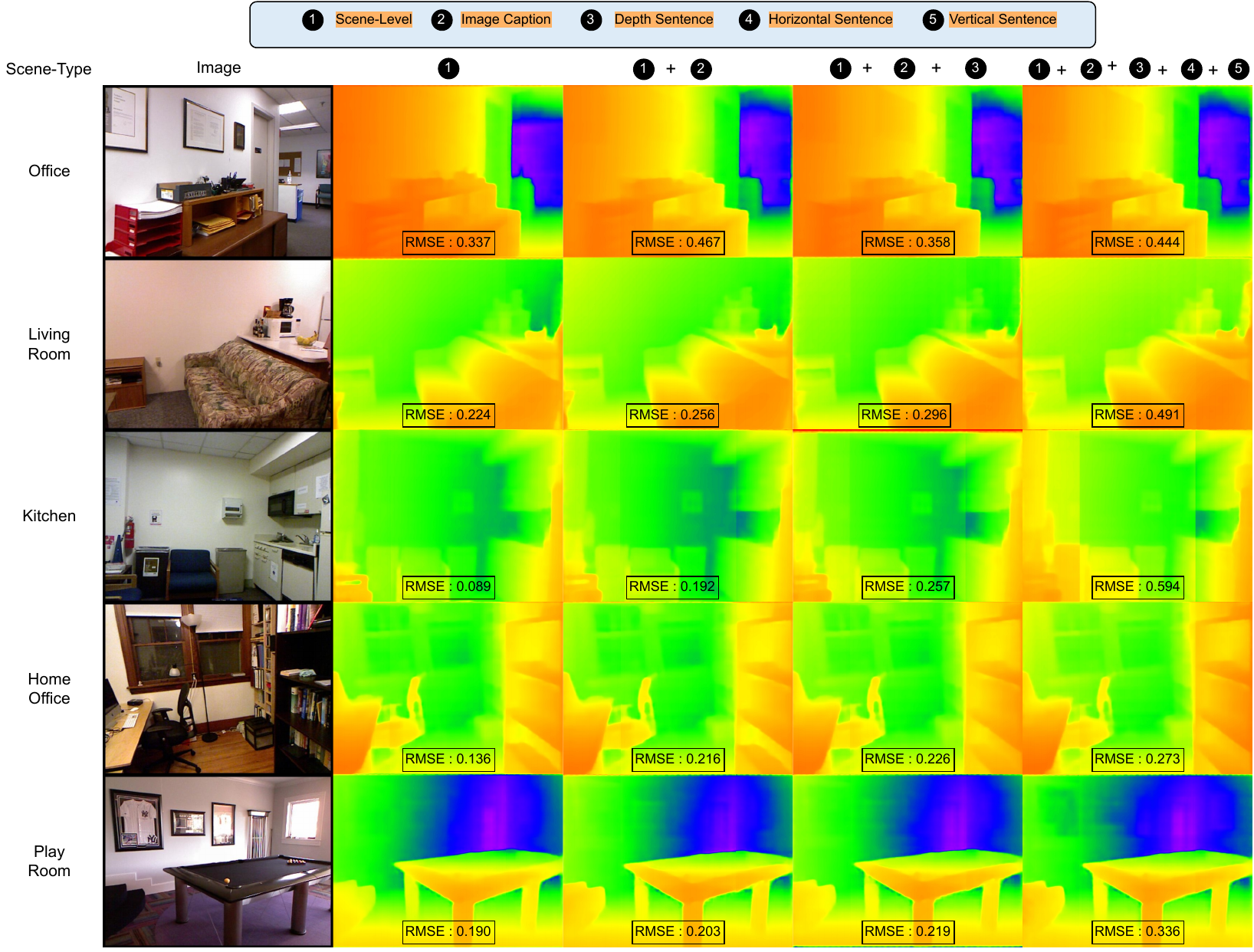}
    \caption{Comparison of Generated Depth Maps, when evaluated in a \textbf{zero-shot} setting, across 5 different scene-types and 4 kinds of natural language guidance. }%
    \label{fig:zero_shot_supp}
\end{figure*}

Figure \ref{fig:supervised_supp} presents additional depth map illustrations as generated by models trained with varying types of natural language guidance. Lastly, in Figure \ref{fig:zero_shot_supp}, we present illustrative results when VPD is evaluated in a zero-shot setting across multiple language modalities.

%% file: tables/room_activity_description.tex
\begin{table}[t]
\centering
\small
\resizebox{\linewidth}{!}{
\begin{tabular}{@{}lp{0.8\linewidth}@{}}
\toprule
Original Scene Name & Activity Level Description \\
\midrule
printer room & room to access and operate printing equipment \\
bathroom & room to attend to personal hygiene and grooming \\
living room & place to relax, socialize, and entertain guests in a house \\
study & room to focus on reading, learning, and researching \\
conference room & room to hold meetings and discussions \\
study room & room to concentrate on academic or professional tasks \\
kitchen & room to prepare and cook meals \\
home office & place to work on professional tasks from home \\
bedroom & room to sleep and rest in a home \\
dinette & place to have informal meals \\
playroom & place to engage in recreational activities and games for kids \\
indoor balcony & place to enjoy views and relax indoors \\
laundry room & room to clean and maintain clothing and fabrics \\
basement & place for storage, recreation, or utilities usually below ground level \\
exercise room & room to workout and engage in physical activities \\
foyer & area of the house to welcome guests and as an entryway \\
home storage & storage area in a house to store items and belongings \\
cafe & place to enjoy beverages and light meals in a social setting \\
furniture store & place to browse and purchase furniture items \\
office kitchen & place to prepare refreshments and snacks in an office \\
student lounge & place to relax and interact in a university or school setting for students \\
dining room & room to have formal meals with family or guests \\
reception room & room to welcome and accommodate visitors \\
computer lab & lab to use computers for learning or work purposes \\
classroom & room to attend educational lectures and lessons \\
office & place to carry out professional tasks and responsibilities \\
bookstore & place to browse and purchase books and literary materials \\
\bottomrule
\end{tabular}
}
\caption{Original scene names in the NYUv2 dataset and their corresponding activity level descriptions.}
\label{tab:activity_level_table}
\end{table}

%% file: tables/ood_class_split.tex
\begin{table}[t]
\centering
\resizebox{\linewidth}{!}{
\begin{tabular}{p{0.6\linewidth}p{0.4\linewidth}}
\toprule
Train  Scenes & Test Scenes \\
\midrule
printer room, bathroom, living room, study, conference room & student lounge, dining room, reception room \\
\midrule 
study room, kitchen, home office, bedroom, dinette, playroom & computer lab, classroom, office, bookstore\\
\midrule 
indoor balcony, laundry room, basement, exercise room & foyer, home storage, cafe, furniture store, office kitchen  \\ 
\bottomrule
\end{tabular}
}
\caption{Modified Train-Test Split of the NYUv2 dataset, as described in the scene distribution supervised setting.}
\label{tab:class_split_ood}
\end{table}